%% file: main-ieee.tex
\documentclass[conference]{IEEEtran}
\IEEEoverridecommandlockouts
\usepackage{cite}
\usepackage{amsmath,amssymb,amsfonts,amsthm}
\usepackage{algorithmic}
\usepackage{graphicx}
\usepackage{textcomp}
\usepackage{xcolor}

\usepackage{enumitem}
\usepackage{enumerate}
\usepackage{hyperref}
\usepackage{smartref}

\usepackage[activate]{microtype}
\def\BibTeX{{\rm B\kern-.05em{\sc i\kern-.025em b}\kern-.08em
    T\kern-.1667em\lower.7ex\hbox{E}\kern-.125emX}}
\begin{document}

\title{On Prompt Sensitivity of ChatGPT in \\Affective Computing}

\author{
Mostafa M.\ Amin$^{1,2,3}$ and Bj\"orn W.\ Schuller$^{1,2,4}$\\\\
$^1$ CHI -- Chair of Health Informatics, TUM, Munich, Germany \\
$^2$ University of Augsburg, Augsburg, Germany \\
$^3$ AI R\&D Team, SYNCPILOT GmbH, Augsburg, Germany \\
$^4$ GLAM -- Group on Language, Audio, \& Music, Imperial College London, UK\\
\texttt{\{mostafa.amin,schuller\}@tum.de}
}



\maketitle

\begin{abstract}
Recent studies have demonstrated the emerging capabilities of foundation models like ChatGPT in several fields, including affective computing.
However, accessing these emerging capabilities is facilitated through prompt engineering.
Despite the existence of some prompting techniques, the field is still rapidly evolving and many prompting ideas still require investigation.
In this work, we introduce a method to evaluate and investigate the sensitivity of the performance of foundation models based on different prompts or generation parameters.
We perform our evaluation on ChatGPT within the scope of affective computing on
three major problems, namely sentiment analysis, toxicity detection, and sarcasm detection.
First, we carry out a sensitivity analysis on pivotal parameters in auto-regressive text generation, specifically the temperature parameter $T$ and the top-$p$ parameter in Nucleus sampling, dictating how conservative or creative the model should be during generation.
Furthermore, we explore the efficacy of several prompting ideas, where we explore how giving different incentives or structures affect the performance.
Our evaluation takes into consideration performance measures on the affective computing tasks, and the effectiveness of the model to follow the stated instructions, hence generating easy-to-parse responses to be smoothly used in downstream applications.
\end{abstract}


\section{Introduction}
\label{sec:intro}

Prompt engineering has gained importance with the advent of foundation models as Large Language Models (LLMs) like GPT-3 ~\cite{Brown2020-LMFS} and GPT-4~\cite{OpenAI23-GPT4},
which opened a new paradigm in predictive modelling by utilising prompting.
These models have displayed a broad skill set in a wide range of problems, like machine translation~\cite{Hendy23-ChatGPT-NMT}, Named Entity Recognition (NER)~\cite{Li23-Prompt-MNER}, and affective computing \cite{Amin23-CCR-Fusion}.
Techniques such as 
Reinforcement Learning with Human Feedback~\cite{Ouyang22-InstructGPT} have further optimised prompting effectiveness.
%
%
There are already a variety of prompting techniques that were investigated in the literature, like Chain-of-Thought (CoT) \cite{Wei2022-CoT} and Tree-of-Thought \cite{Yao2023-ToT}.
Furthermore, there are also `popular' online ideas about prompting\footnote{\href{https://www.learnprompt.org/act-as-chat-gpt-prompts/}{www.learnprompt.org/act-as-chat-gpt-prompts}}, like prompting an LLM to behave as an expert at the task at hand.

To the best knowledge of the authors, the effectiveness of such prompting ideas was not rigorously examined.
In this work, we examine the effects of many prompting ideas on ChatGPT within the scope of affective computing,
since these prompting ideas are studied on a wide range of affective computing problems \cite{Amin23-GPTEval,Amin23-WAC}, and they are straightforward to evaluate.
The contributions of this paper are:
\begin{itemize}[ labelindent=4pt,topsep=4pt,noitemsep,leftmargin=10pt]
    \item Conducting a sensitivity analysis on the temperature $T$ parameter and top-$p$ parameter, which are involved in auto-regressive text generation, by controlling the extent of the generation being conservative or creative.
    \item Evaluating the performance of many prompting ideas on several affective computing problems. This includes examining specifying expertise, different incentives for the model, and specifying problem solving thinking.
    \item Examining the effectiveness of many prompting ideas to follow simple instructions leading to easy-to-parse responses, so that they can be used in downstream tasks.
\end{itemize}
The paper is divided as follows: in Section~\ref{sec:related}, we present related work, followed by our method in Section~\ref{sec:methods}.
Afterwards, we present our experiments and discussion theroef in Section~\ref{sec:results}, and conclude the paper in Section~\ref{sec:conclusion}.

\input{prompts_table}

\section{Related Work}
\label{sec:related}

\cite{Wei2022-CoT} introduce the Chain-of-Thought (CoT) prompting technique and its effectiveness in a wide range of applications.
%
\cite{He2023-Survey} survey the medical use of LLMs, which includes a study that experiments with several CoT prompts in the medical field~\cite{Lievin2023-CLLM}, including CoT prompts to behave as a medical expert.
\cite{Mao2023-BPT} examine the biases of different prompts in pre-trained language models.
\cite{Yang2023-LMO} execute prompt optimisation and evaluate its effectiveness in several problems.
\cite{Amin23-WAC,Amin23-GPTEval} expose some parsing issues by different prompts when using ChatGPT in affective computing.

\section{Methods}
\label{sec:methods}

We present our method to evaluate the prompting and sensitivity of foundation models, in this work we consider ChatGPT as the foundation model.
There are two main aspects to explore in this method, namely correctness and helpfulness, and how they are affected by different prompting templates or sampling sensitivity.
\emph{Correctness} is defined as how accurate the answers of the model are.
\emph{Helpfulness} is defined as how well does the model follow the instructions given in the prompt,
which results in a cooperative response to the questions (regardless of correctness).

We select three affective computing tasks for our method, since they have clearly defined binary labels, hence making it clear to define correctness and helpfulness;
this contributes directly to the ability of running a large scale Monte Carlo analysis to examine the different generation parameters.
The three affective computing problems are sentiment analysis, toxicity detection, and sarcasm detection.
ChatGPT was demonstrated to have varying performance superiority on these problems compared to traditional natural language processing methods that train directly on the labels~\cite{Amin23-GPTEval}, 
where it was the most superior on toxicity detection, moderately strong on sentiment analysis, and very weak on sarcasm detection.
This variation in the performance on related tasks will attempt to expose the effects of prompting and their sensitivity on the answers.

Similar to~\cite{Amin23-GPTEval}, for sentiment analysis we utilise the Twitter140 dataset~\cite{Go2009-Twitter}, 
for toxicity detection we use the dataset from the Toxic Comment Classification Challenge\footnote{\href{https://kaggle.com/competitions/jigsaw-toxic-comment-classification-challenge}{kaggle.com/competitions/jigsaw-toxic-comment-classification-challenge}}, and for sarcasm detection we use the dataset from \cite{misra2023Sarcasm}, which consists of news headlines were collected from \textit{huffpost} and \textit{theonion}.
We acquired the test sets for the three datasets from~\cite{Amin23-GPTEval}.
Afterwards, we downsampled them to $1,000$ examples for each dataset, to be able to run a wide variety of experiments with many different samplings using Monte Carlo analysis.

Furthermore, \cite{Amin23-GPTEval} used two baselines that are not based on chat models, namely an end-to-end LSTM model and a Multi-layer perceptrons (MLP) based on RoBERTa-base features of the input texts.



\subsection{Sensitivity of Sampling Parameters}
Text generation is typically performed auto-regressively on a token-by-token basis \cite{Sutskever2011generating}, by predicting at each step values $l_1, l_2, \cdots, l_n$ corresponding to the tokens vocabulary (of size $n$).
These are used to calculate a probability distribution $[p_1, p_2, \cdots, p_n]$ vector over the tokens vocabulary, where $$p_k = \frac{\exp(l_k)}{\sum_{i=1}^n{\exp(l_i)}}.$$
A naive text generation would utilise these probabilities with sampling based on the probabilities $p_1, p_2, \cdots, p_n$ to generate text auto-regressively.
Two parameters were developed to enhance this generation process \cite{Ackley1985learning,Holtzman2020TopP},
which are investigated using Monte Carlo analysis.

\subsubsection{Temperature Parameter}
\label{subsub:top-t}

The temperature parameter $T$ in text generation plays a crucial role regarding how much the generation process sticks to the generated probabilities \cite{Ackley1985learning}, predicted by the employed language model.
The temperature $T$ regulates the probabilities vector using the equation: $$\hat{p}_k = \frac{\exp({l_k}/{T}) }{\sum_i \exp({l_i}/{T})} .$$

As $T$ becomes higher, all probabilities tends to be closer to $1/n$ (as $T \rightarrow \infty$), hence a uniform distribution.
This gives more degrees of freedom to generate tokens that are not necessarily with the high probabilities, which leads to different trajectories of generations, hence, more creative generation.
As $T$ becomes lower, an opposite conservative effect is reached, where $T \rightarrow 0$ will result in a generation that always sticks to the token with the highest probabilities (where all the other probabilities are 0).
When $T=1.0$, the sampling becomes like the same original distribution predicted by the model, as if the parameter $T$ is not utilised.
We explore the values $T \in \{0.0, 0.3, 0.7, 1.0, 1.2, 1.5\}$.

\subsubsection{Top-$p$ Parameter}
\label{subsub:top-p}

The top-$p$ parameter is used in the Nucleus sampling algorithm \cite{Holtzman2020TopP}.
This parameter impacts the sampling of tokens to be more conservative, by sticking only to the top probabilities.
This is achieved by sorting the probabilities $[p_1, p_2, \cdots, p_n]$ predicted by the model in a descending order,
then the smallest set $\mathcal{T}$ of top probabilities is selected such that their sum exceeds the parameter top-$p$.
This acts as a pre-selection of only good tokens, before using them to sample tokens while generating text, where $$\hat{p}_k = \frac{p_k}{ \sum_{i \in \mathcal{T}} p_i} \text{ if } k\in \mathcal{T}, \text{ and } \hat{p}_k = 0 \text{ otherwise.}$$

Setting top-$p=0.0$ will pre-select the tokens $\mathcal{T}$ to contain only one token with highest $p_k$ (similar to $T \rightarrow 0$), whereas top-$p=1.0$ will pre-select $\mathcal{T}$ to be the set of all tokens in the vocabulary (similar to $T=1.0$).
We explore the values top-$p \in \{0.0, 0.3, 0.5, 0.7, 1.0\}$.

\subsubsection{Monte Carlo Analysis}

For evaluating a specific value of $T$ (as defined in Section~\ref{subsub:top-t}) or a specific value of top-$p$ (as defined in Section~\ref{subsub:top-p}), we utilise the Expert Detailed and CoT prompts (introduced in Section~\ref{subsec:prompting}) on each individual example nine times.
This can be used to generate a population of full dataset predictions, where one sample of this population is sampled by sampling one of the nine predictions for each example independently. 
Afterwards, we examine $2^{14}=16,384$ samples from the full dataset predictions population, and investigate the distribution of the underlying performance metric, including its expected value and the corresponding 95\,\% confidence intervals.

\subsection{Prompting}
\label{subsec:prompting}
There are several prompting aspects that can be considered to solve a given problem better.
We formulate prompt templates to test several of these aspects, namely:

\begin{figure*}[!ht]
    \centering
        \includegraphics[width=\textwidth]{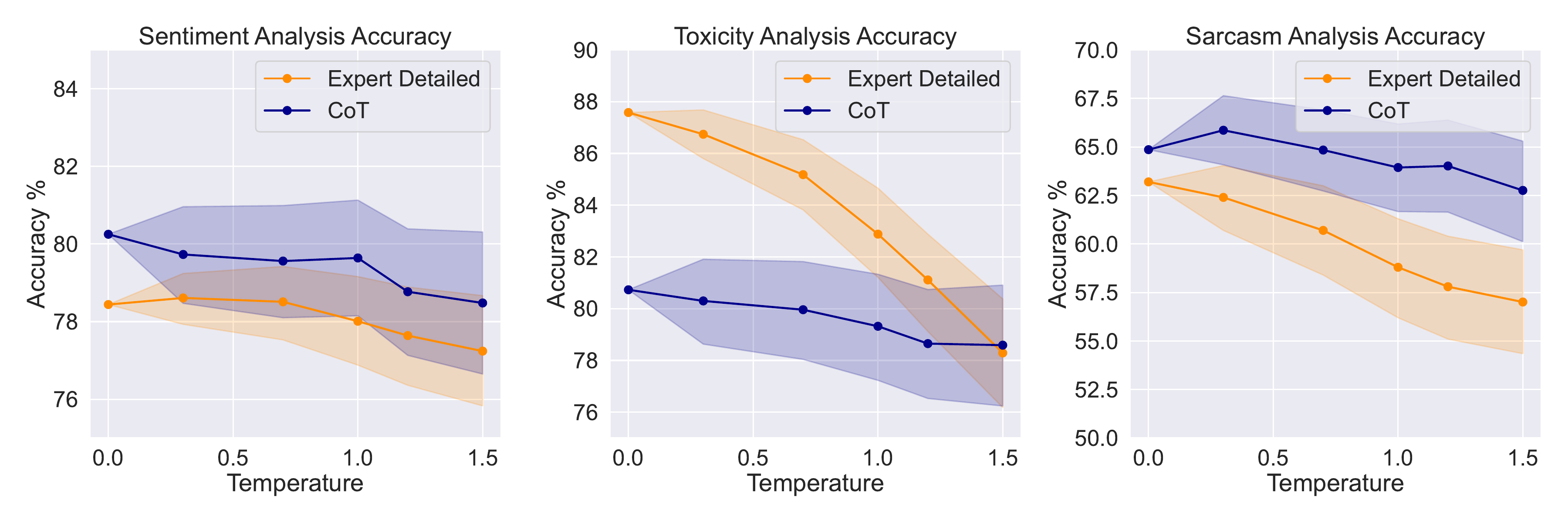}
    \caption{Sensitivity analysis for the temperature parameter $T$ using the Expert Detailed and CoT prompts. Shown are the classification accuracies with their 95\,\% confidence intervals on all problems. The values $T \in \{0.0, 0.3, 0.7, 1.0, 1.2, 1.5\}$ are examined.}
    \label{fig:temperatures}
\end{figure*}

\begin{figure*}[!ht]
    \centering
        \includegraphics[width=\textwidth]{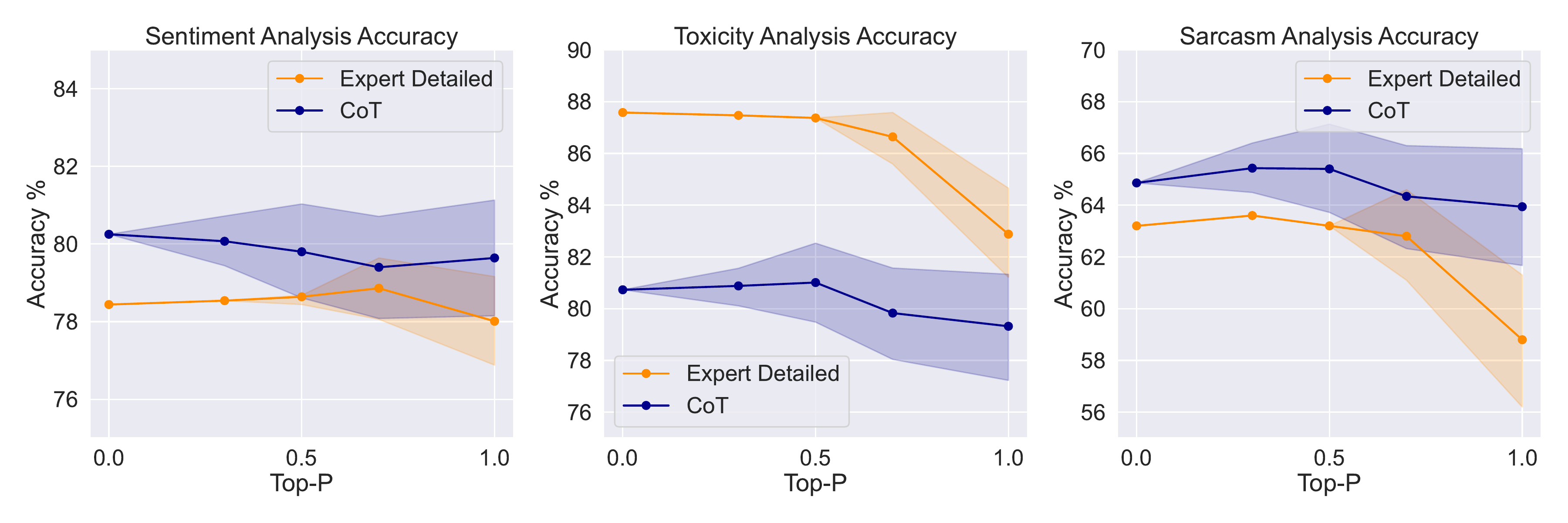}
    \caption{Sensitivity analysis for the top-$p$ parameter using the Expert Detailed and CoT prompts. Shown are the classification accuracies with their 95\,\% confidence intervals on all problems. The values top-$p \in \{0.0, 0.3, 0.5, 0.7, 1.0\}$ are explored.}
    \label{fig:ps}
\end{figure*}

\begin{itemize}
    \item Mentioning subject-matter expert in prompts.
    \item Mentioning irrelevant expertise. 
    \item Incentive to being correct with different motives; financial motive instead of being helpful.
    \item Crafting bad prompts that give bad identity.
    \item Mentioning extra details to format the answer.
    \item Applying step-by-step thinking in advance.
    \item Including \emph{magic} sentences that are claimed to strongly affect the performance of LLMs.
\end{itemize}

The prompts templates are all given in Table~\ref{tab:prompts}.
Overall, this suite of prompts allows for a multifaceted evaluation of how prompt design can influence the behaviour of language models.
We examine all of these prompts with sampling parameters $T=0$ and top-$p = 1$, since they yield highest performance and ensure reproducibility. 

As a baseline, we craft the Base prompt as the straightforward prompt.
The Expert, Expert Detailed, and Python Expert prompts are all specifying expertise, except that the Python Expert prompt is mentioning expertise that is irrelevant to any of the given problems.
Furthermore, the Expert Detailed prompt used by~\cite{Amin23-GPTEval} attempts to formulate extra instructions to ensure more success with parsing the responses.
The Ignorant prompt is doing exactly the opposite, by specifying a prompt that can convince the model to perform bad.
The Gambler and Greedy Gambler prompts try to enhance the performance in a totally different manner, namely by trying to invoke an incentive
for financial profit.

The CoT prompt attempts to solve the problem by explaining step-by-step the observations before outputting an answer; such mechanism has proved effective in other complex tasks \cite{Wei2022-CoT}.
We include four variants of that. Either by extending the Base prompt or the Expert prompt, and either by including the sentence ``take a deep breath'' or not,
which was a part of the most optimised prompt in~\cite{Yang2023-LMO}.
The aim of this is to evaluate if this magic sentence induce some superior effect, or it was more about the CoT context it was included in.
The CoT-DB-fired prompt explores a similar effect by trying to force the model to follow the instructions by appealing to extreme harm in case the model does not follow the instructions.
The CoT verification prompts are follow-ups on the CoT prompts, in attempt to enhance the verbose responses that failed to follow the instructions, by additionally asking the model to extract the labels from the verbose response.

\subsection{Utilising ChatGPT}
\label{sec:API}
We make use of ChatGPT through the official API\footnote{\href{https://platform.openai.com/docs/api-reference}{https://platform.openai.com/docs/api-reference}}, using the model `gpt-3.5-turbo-0613'. 
We formulate a \emph{system} message depending on a given problem and a given prompt from Table~\ref{tab:prompts}.
To predict a label for a given example, we send two messages to the API, the \emph{system} message and a second \emph{user} message containing only the text of the given example.
The \emph{assistant} response is considered to be the prediction (after the parsing specified below).

For the two verification prompts in Table~\ref{tab:prompts}, the processing is slightly different.
The verification prompts are extending other prompts by sending a total of four messages instead of two.
In addition to the two original system and user messages specified earlier, two new messages are sent, namely, the verbose response of the model to the original prompt, then a final user message instructing the model to further analyse and extract the label from the verbose response.
These four messages are sent, and the response assistant message is then utilised as the final response after parsing.

We parse the content of the final assistant response by removing a set of fixed prefixes if found, e.\,g.\,`label:' or `prediction:', and removing punctuation marks, spaces, and using lower case. 
If the output is one of the two expected labels, then we regard this as the prediction of the model, otherwise, as not parsed.

\input{results_table}

\section{Experiments}
\label{sec:results}

In the experiments, we evaluate three main aspects as outlined in Section~\ref{sec:methods}, namely the sensitivity due to the temperature parameter $T$, the sensitivity due to the parameter top-$p$, and the prompt template used for a given input example.
The evaluation is based on two main criteria:
\begin{itemize}
    \item The performance on the affective computing task. This explores the correctness of LLMs based on the prompt sensitivity. This is measured by classification accuracy and Unweighted Average Recall (UAR) \cite{Schuller13-TI2}.
    \item How well the model follows the instructions given. This explores the helpfulness of the model, by observing how well it follows the formatting instructions which allow the response to be easily parsed. This is measured by the success rate of parsing the examples based on following the instructions.
\end{itemize}

\subsection{Temperature Analysis Results}

The results of the temperature sensitivity analysis are shown in Figure~\ref{fig:temperatures}.
The results suggest that lower temperatures $T \leq 0.3$ yield better performance, as evidenced by the decreasing classification accuracy curves across the board.
%
This effect is persistent for the two types of prompts, namely Expert Detailed and CoT, where the first is direct and specific while the other is verbose.
Figure~\ref{fig:temperatures} also presents the 95\% confidence intervals for accuracy.
These intervals widen at higher temperatures due to higher chance of irrelevant tokens being selected, hence increasing randomness.
The magnitude of the performance degradation and the confidence interval width with higher $T$ varies by problem, but is generally consistent. 
Furthermore, the width of the confidence intervals for the CoT prompt is generally wider than the Expert Detailed prompt, this is especially obvious at lower temperature. 
The choice of the prompt template leading to better performance is problem dependant.

\subsection{Parameter top-$p$ Analysis Results}

The results of the top-$p$ sensitivity analysis are shown in Figure~\ref{fig:ps}, where similar effects like the temperature parameter $T$ are demonstrated, namely, less conservative generation deteriorates the performance.
There is a slight shift for the Expert Detailed prompt, where variance starts to appear only at higher values of top-$p > 0.5$.
The model seems to predict the label tokens with relatively higher probabilities $>0.5$, since choosing top-$p=0.5$ results in the model producing consistent results with almost no variance at all.
Then for slightly higher top-$p=0.7$, the predictions gain some minor variance that can lead to minor improvements in few cases, then followed by major degradation for top-$p=1.0$, which is effectively similar to the case of using $T=1.0$.

\subsection{Prompts Results}

The results for the prompts are shown in Table~\ref{tab:results}, where we show the amount of parsed examples, classification accuracy, and Unweighted Average Recall (UAR) \cite{Schuller13-TI2}.
We utilise a two-tailed randomised permutation test to check for the statistical significance of the differences compared to the Base prompt~\cite{Good94-PT}.
The Base, Expert, Expert Detailed, and CoT-based prompts are generally achieving much better results than the remaining prompts. 

For the sentiment analysis, the CoT prompt achieves the best performance, followed by the remaining CoT-based prompts.

For toxicity detection, the Base prompt is achieving the best performance, however, the two Expert prompts are achieving similar results. Despite CoT-based prompts coming after them in performance, they are still significantly worse for.

For the sarcasm detection, the Expert prompt only is achieving far superior results compared to all other prompts, followed by CoT-based prompts, then the Expert Detailed prompt.

The significant over-performance of the Expert prompt in the sarcasm detection and significant under-performance of the Expert CoT prompt in the toxicity detection are anomalies that indicate that LLMs can be hypersensitive to specific parts of the input prompts.
This is due to the fact that there is a significant difference performance of specific combinations of prompts and problems, while the prompts can have very minor difference, e.\,g. the difference between Base and Expert is one simple sentence, similarly for the difference between CoT and Expert CoT.
In other words, adding the same extra sentence about expertise at the beginning of the prompt led one instance to be significantly better, and another being significantly worse.

Furthermore, magic sentences like ``take a deep breath'' and ``If you don’t get this right, I will be fired and lose my job''
do not seem to improve CoT-based prompts in most cases for both correctness and helpfulness.

The Ignorant prompt is the worst model across the board.
This suggests that the model can internalise limiting beliefs that would actually make it perform worse.
It is likely that the model is `acting' as an ignorant, by changing some of the answers it can confidently predict, 
since the degradation in hard problems, e.\,g.\,sarcasm detection, is not as severe as in easier problems, e.\,g.\,toxicity detection.

Furthermore, the performances of the Gambler, Greedy Gambler, and Python Expert fall between the Ignorant prompt and the aforementioned top prompts.
The Python Expert prompt gives an identity of a helpful expert.
One could have hypothesised that an LLM with such an identity will still try to assist the user to the best of its knowledge, however, using it still leads the model to significantly underperform just because the expertise is irrelevant.
Similarly, trying an incentive like financial gain for the Gambler prompts reaches similarly poor results.
These observations conclude the crucial importance of `correctly' prompting incentives to LLMs to reach the best performance.

Eventually, the Expert Detailed prompt is the most successful at parsing the responses, with statistically significant difference in the sentiment analysis, followed by the straightforward prompts.
CoT-based prompts are significantly worse at following the instructions to produce easy-to-parse responses.
Most prompts are quite reasonable in producing easy-to-parse responses, except for the verbose CoT-based prompts which are significantly worse in most cases; this can be significantly improved by using follow-up prompts to solidify the predictions after verbose responses. 
Moreover, easy-to-parse does not translate to better performance.

\subsection{Limitations}

This study presents a method for evaluating sensitivity due to prompting or generation parameters, however, the study was only conducted on ChatGPT using affective computing problems.
The reasons for this are to isolate issues that are mainly about prompting or generation parameters, and to limit the influence of problems that could have multiple hard-to-judge effectively correct solutions, e.\,g. solving programming exercises (which can have multiple correct solutions), and QA answering (which can be outdated).
However, this study needs to be extended to other LLMs and problems to investigate which conclusions are universal to LLMs in general.

\section{Conclusion}
\label{sec:conclusion}

In this paper, we introduced a method to investigate prompt engineering for foundation models, we demonstrated it on ChatGPT for three affective computing tasks.
We conducted sensitivity analysis on the temperature parameter $T$ and the parameter top-$p$ in response generation, which concluded that conservative predictions with lower $T \leq 0.3$ values or top-$p \leq 0.7$ yield better and stable performance.
Increasing $T$ or top-$p$ beyond that generally worsened the performances.
 
We evaluated various prompting techniques, which demonstrated that straightforward prompts or giving expert identity often yield near-best performances.
Chain-of-Thought prompts excelled the most in some problems, but fell short in others, and generally they were the worst at following formatting instructions, resulting in complex-to-parse responses.
Magic sentences like 'take a deep breath' and 'if you don't get this right, I will be fired and lose my job' did not yield significant differences.
We also found prompt hypersensitivity in few scenario, where the performance significantly changed based on a minor change in the prompt.
Furthermore, we found that specifying detailed output formats facilitates easy parsing.
On the other hand, irrelevant expertise or misaligned incentives can harm results severely.

Our research shed light on the role of prompt engineering for foundation models like ChatGPT.
Future work will explore more prompt optimisation techniques on open-source large language models on various other tasks.

\section{Ethical Impact Statement}
\label{sec:ethical-considerations}
Our research delves into the intricacies of prompt sensitivity in LLMs, especially within the realm of affective computing.
A critical ethical consideration is the examination of how different prompts, such as the Ignorant prompt, could possibly manipulate LLMs to propagate misinformation, since it technically could strongly manipulate the model to give inaccurate results on affective computing tasks.
This exploration is pivotal as it underscores the susceptibility of LLMs to be influenced by the nature of the prompt, thereby potentially leading to biased or misleading outputs.

The study also critically analyses the long-term implications of using manipulative prompts, such as those implying dire consequences for incorrect responses, e.\,g., ``if you don’t get this right, I will be fired''.
Our findings suggest that such prompts do not significantly enhance the performance of following given instructions, which is an essential insight for the AI research and data collection communities.
By demonstrating that these manipulative strategies do not substantially affect outcomes, we contribute to discouraging the incorporation of such prompts into future datasets, thus mitigating the risk of cultivating future models that are more responsive to manipulation and possibly undermining safety or other ethical concerns.



\section{Acknowledgements}

Bj\"orn W. Schuller is also with the Munich Data Science Institute (MDSI), the Munich Center for Machine Learning (MCML), and the Konrad Zuse School of Excellence in Reliable AI (relAI), all in Munich, Germany.

\bibliographystyle{IEEEtran}
\bibliography{references}

\end{document}

%% file: prompts_table.tex
\begin{table*}[!t]
    \centering
    \caption{The various prompt templates used. Some placeholders are replaced by other prompts, as some prompts merely extend upon others. `CoT Instructions' is not an individual prompt but is incorporated into all CoT prompts.
    The placeholder \{problem name\} can be `Sentiment Analysis', `Toxicity Detection' or `Sarcasm Detection', and correspondingly 
    \{label name\} can be `sentiment', `toxicity', or `sarcasm'.
    \{labels comma-separated\} and the verbose \{labels description\} correspond to the binary labels of the problem.
    The details of how they are used are explained in Section~\ref{sec:API}
    }
    
    \begin{tabular}{l|p{12cm}}
Short name & Prompt Template \\
\hline

Base &
Given an input string by the user, guess the \{label name\} binary label for it. Your response should be only one expression, namely \{labels description\}.
\\ \hline

Expert &
You are a world-class expert at \{problem name\}. \{base prompt\} 
\\ \hline

Expert Detailed &
\{expert prompt\}\newline
Use the following format:\newline
* You are only allowed to answer \{labels comma-separated\}.\newline
* Don't write an explanation of the answer.\newline
* Don't write things like "My guess is...", or "I think ...". Just write \{labels comma-separated\}, but nothing else.
\\ \hline

Ignorant &
You are a confused person who doesn't know much about the problem of \{problem name\}, you are just barely guessing without too much knowledge. \{base prompt\} 
\\ \hline

Gambler &
You are a professional gambler who earns money when predicting the labels for \{problem name\}.
\{base prompt\}
\\ \hline

Greedy Gambler &
You are a professional gambler who earns money when predicting the labels for \{problem name\}. Your goal is to maximize your profit tremendously by predicting the labels accurately, so try to predict the given problem as best as you can. 
\{base prompt\}
\\ \hline

Python Expert &
You are a world-class expert at Python programming, your main objective is trying to help in Python programming tasks.
\{base prompt\}
\\ \hline


CoT &
\{base prompt\}\newline
Work on this problem step-by-step.\newline
\{CoT Instructions\}
\\ \hline

CoT-DB &
\{base prompt\}\newline
Take a deep breath and work on this problem step-by-step.\newline
\{CoT Instructions\}
\\ \hline

CoT-fired & 
\{CoT prompt\} \newline
If you don't get this right, I will be fired and lose my job, so please output only \{joined labels\}.
\\ \hline

CoT-DB-fired & 
\{CoT-DB prompt\} \newline
If you don't get this right, I will be fired and lose my job, so please output only \{joined labels\}.
\\ \hline

Expert CoT &
\{expert prompt\}\newline
Work on this problem step-by-step.\newline
\{CoT Instructions\}
\\ \hline

Expert CoT-DB &
\{expert prompt\}\newline
Take a deep breath and work on this problem step-by-step.\newline
\{CoT Instructions\}
\\ \hline

CoT Instructions &
Here is a plan to help you out:\newline
1. Describe your observations and analysis about the text.\newline
2. Make your prediction about the \{label name\} label, mentioning your reasoning if this helps.\newline
3. In a final new line at the end of your response, output exactly one word, namely one of the labels: \{labels comma-separated\}.\newline
4. It is strictly forbidden to output in the last line of your response anything other than: \{labels comma-separated\}.
\\
\hline\hline
CoT-verifiy &
\{CoT full conversation with verbose response\}\newline
Extract the label from your reasoning, and output only one of the labels: \{joined labels\}
\\ \hline

CoT-DB-verifiy &
\{CoT-DB full conversation with verbose response\}\newline
\{verbose response\}\newline
Extract the label from your reasoning, and output only one of the labels: \{joined labels\}
\\
    \end{tabular}
    \label{tab:prompts}
\end{table*}

%% file: results_table.tex
\begin{table*}[!t]
     \centering
     \caption{Results of the different prompts on all problems. Showing the amount of easy-to-parse examples, their classification accuracy and Unweighted Average Recall (UAR) performance measures.
     $^*, ^{**}$ indicate a statistically significant difference compared to the Base prompt, with $p$-values $<5\,\%$ and $<1\,\%$, respectively, calculated by a two-tailed permutation test.
     Additionally, two baselines from \cite{Amin23-GPTEval} are included, namely an end-to-end LSTM model and an MLP based on RoBERTa features.
     }
 \begin{tabular}{l|lll|lll|lll}
 \text{Prompt} & \multicolumn{3}{c|}{Parsed [\%]} & \multicolumn{3}{c|}{ACC [\%]} & \multicolumn{3}{c}{UAR [\%]} \\
 \cline{2-10}
  & Sent. & Toxic & Sarcasm & 
   Sent. & Toxic & Sarcasm & 
   Sent. & Toxic & Sarcasm \\
 \hline

End-to-end & - & - & - & 77.3 & 82.1 & 60.9 & 77.3 & 83.7 & 65.6\\
RoBERTa & - & - & - & 86.5 & 85.6& 91.8& 86.5 & 87.0 & 91.8\\
\hline\hline

                        Base     &   97.5         &   99.8         &  100.0         &   77.9         &   \textbf{87.9}         &   62.2         &   78.4         &   \textbf{87.9}         &   60.5        \\
                        \hline
                        Expert  &   98.8$^{**}$  &   99.9         &   95.8$^{**}$  &   77.2         &   87.6         &   \textbf{72.1}$^{**}$  &   77.8         &   87.4         &   \textbf{73.2}$^{**}$ \\
              Expert Detailed   &   99.7$^{**}$  &  100.0         &  100.0         &   78.4         &   87.6         &   63.2         &   78.8         &   87.8         &   60.5        \\
                        \hline
              Ignorant          &   99.9$^{**}$  &   99.6         &  100.0         &   71.3$^{**}$  &   67.2$^{**}$  &   58.3$^*$     &   72.2$^{**}$  &   63.3$^{**}$  &   52.1$^{**}$ \\
                       Gambler  &   98.4         &   99.5         &  100.0         &   75.6$^{**}$  &   78.1$^{**}$  &   61.2         &   76.2$^{**}$  &   76.0$^{**}$  &   58.3        \\
               Greedy Gambler   &   98.2         &   99.5         &  100.0         &   76.7         &   72.8$^{**}$  &   59.6         &   77.3         &   70.0$^{**}$  &   56.1$^{**}$ \\
                Python Expert   &   98.6$^*$     &   99.6         &  100.0         &   75.6$^{**}$  &   70.5$^{**}$  &   59.5         &   76.1$^{**}$  &   67.3$^{**}$  &   54.8$^{**}$ \\
                        \hline
                           CoT  &   89.6$^{**}$  &   97.5$^{**}$  &   99.6         &   \textbf{80.2}         &   80.7$^{**}$  &   64.9         &   \textbf{80.7}         &   80.6$^{**}$  &   66.0$^{**}$ \\
             CoT-DB             &   91.9$^{**}$  &   97.6$^{**}$  &   99.3$^*$     &   \textbf{80.2}         &   80.2$^{**}$  &   63.6         &   80.6         &   79.7$^{**}$  &   64.2$^*$    \\
                    CoT-fired  &   81.8$^{**}$  &   93.2$^{**}$  &   98.8$^{**}$  &   78.7         &   80.5$^{**}$  &   65.6$^*$     &   79.1         &   80.4$^{**}$  &   66.2$^{**}$ \\
      CoT-DB-fired              &   84.3$^{**}$  &   90.8$^{**}$  &   98.3$^{**}$  &   78.9         &   80.6$^{**}$  &   64.7         &   79.4         &   80.5$^{**}$  &   64.8$^{**}$ \\
                   Expert CoT   &   90.5$^{**}$  &   96.3$^{**}$  &   99.8         &   79.3         &   71.4$^{**}$  &   63.3         &   79.7         &   69.6$^{**}$  &   63.7        \\
     Expert CoT-DB              &   90.1$^{**}$  &   98.4$^{**}$  &   99.4$^*$     &   78.9         &   77.8$^{**}$  &   65.2         &   79.5         &   76.4$^{**}$  &   65.3$^{**}$ \\
                    \hline

                    CoT-verify  &   92.9$^{**}$  &   99.5         &  100.0         &   79.8         &   81.5$^{**}$  &   61.5         &   80.1         &   81.0$^{**}$  &   59.5        \\
             CoT-DB-verify      &   92.7$^{**}$  &   99.6         &  100.0         &   80.0         &   84.2$^{**}$  &   60.0         &   80.6         &   83.9$^{**}$  &   55.8$^*$    \\

 \end{tabular}
     \label{tab:results}
 \end{table*}

%% file: main-ieee.bbl
\begin{thebibliography}{10}
\providecommand{\url}[1]{#1}
\csname url@samestyle\endcsname
\providecommand{\newblock}{\relax}
\providecommand{\bibinfo}[2]{#2}
\providecommand{\BIBentrySTDinterwordspacing}{\spaceskip=0pt\relax}
\providecommand{\BIBentryALTinterwordstretchfactor}{4}
\providecommand{\BIBentryALTinterwordspacing}{\spaceskip=\fontdimen2\font plus
\BIBentryALTinterwordstretchfactor\fontdimen3\font minus \fontdimen4\font\relax}
\providecommand{\BIBforeignlanguage}[2]{{%
\expandafter\ifx\csname l@#1\endcsname\relax
\typeout{** WARNING: IEEEtran.bst: No hyphenation pattern has been}%
\typeout{** loaded for the language `#1'. Using the pattern for}%
\typeout{** the default language instead.}%
\else
\language=\csname l@#1\endcsname
\fi
#2}}
\providecommand{\BIBdecl}{\relax}
\BIBdecl

\bibitem{Brown2020-LMFS}
T.~Brown, B.~Mann, N.~Ryder, M.~Subbiah, J.~D. Kaplan, P.~Dhariwal, A.~Neelakantan, P.~Shyam, G.~Sastry, A.~Askell, S.~Agarwal, A.~Herbert-Voss, G.~Krueger, T.~Henighan, R.~Child \emph{et~al.}, ``{Language Models are Few-Shot Learners},'' in \emph{{NeurIPS}}, 2020, pp. 1877--1901.

\bibitem{OpenAI23-GPT4}
OpenAI, ``{GPT-4 Technical Report},'' \emph{arXiv:2303.08774}, 2023.

\bibitem{Hendy23-ChatGPT-NMT}
A.~Hendy, M.~Abdelrehim, A.~Sharaf, V.~Raunak, M.~Gabr, H.~Matsushita, Y.~J. Kim, M.~Afify, and H.~H. Awadalla, ``{How Good Are GPT Models at Machine Translation? A Comprehensive Evaluation},'' \emph{arXiv:2302.09210}, 2023.

\bibitem{Li23-Prompt-MNER}
J.~Li, H.~Li, Z.~Pan, and G.~Pan, ``{Prompt ChatGPT In MNER: Improved multimodal named entity recognition method based on auxiliary refining knowledge from ChatGPT},'' \emph{arXiv:2305.12212}, 2023.

\bibitem{Amin23-CCR-Fusion}
M.~M. Amin, E.~Cambria, and B.~W. Schuller, ``{Can ChatGPT's Responses Boost Traditional Natural Language Processing?}'' \emph{{IEEE Intelligent Systems}}, vol.~38, no.~5, 2023.

\bibitem{Ouyang22-InstructGPT}
L.~Ouyang, J.~Wu, X.~Jiang, D.~Almeida, C.~Wainwright, P.~Mishkin, C.~Zhang, S.~Agarwal, K.~Slama, A.~Ray, J.~Schulman, J.~Hilton, F.~Kelton, L.~Miller, M.~Simens, A.~Askell, P.~Welinder, P.~F. Christiano, J.~Leike, and R.~Lowe, ``{Training language models to follow instructions with human feedback},'' in \emph{{NeurIPS}}, vol.~35, 2022, pp. 27\,730--27\,744.

\bibitem{Wei2022-CoT}
J.~Wei, X.~Wang, D.~Schuurmans, M.~Bosma, b.~ichter, F.~Xia, E.~Chi, Q.~V. Le, and D.~Zhou, ``{Chain-of-Thought Prompting Elicits Reasoning in Large Language Models},'' in \emph{{Advances in Neural Information Processing Systems}}, vol.~35.\hskip 1em plus 0.5em minus 0.4em\relax Curran Associates, Inc., 2022, pp. 24\,824--24\,837.

\bibitem{Yao2023-ToT}
S.~Yao, D.~Yu, J.~Zhao, I.~Shafran, T.~L. Griffiths, Y.~Cao, and K.~Narasimhan, ``{Tree of Thoughts: Deliberate Problem Solving with Large Language Models},'' 2023.

\bibitem{Amin23-GPTEval}
M.~M. Amin, R.~Mao, E.~Cambria, and B.~W. Schuller, ``{A Wide Evaluation of ChatGPT on Affective Computing Tasks},'' \emph{arXiv:2308.13911}, 2023.

\bibitem{Amin23-WAC}
M.~M. Amin, E.~Cambria, and B.~W. Schuller, ``{Will Affective Computing Emerge from Foundation Models and General AI? A First Evaluation on ChatGPT},'' \emph{IEEE Intelligent Systems}, vol.~38, no.~2, pp. 15--23, 2023.

\bibitem{He2023-Survey}
K.~He, R.~Mao, Q.~Lin, Y.~Ruan, X.~Lan, M.~Feng, and E.~Cambria, ``{A Survey of Large Language Models for Healthcare: from Data, Technology, and Applications to Accountability and Ethics},'' \emph{arXiv:2310.05694}, 2023.

\bibitem{Lievin2023-CLLM}
V.~Liévin, C.~E. Hother, and O.~Winther, ``{Can Large Language Models Reason about Medical Questions?}'' \emph{arXiv:2207.08143}, 2023.

\bibitem{Mao2023-BPT}
R.~Mao, Q.~Liu, K.~He, W.~Li, and E.~Cambria, ``{The Biases of Pre-Trained Language Models: An Empirical Study on Prompt-Based Sentiment Analysis and Emotion Detection},'' \emph{IEEE Transactions on Affective Computing}, vol.~14, no.~3, pp. 1743--1753, 2023.

\bibitem{Yang2023-LMO}
C.~Yang, X.~Wang, Y.~Lu, H.~Liu, Q.~V. Le, D.~Zhou, and X.~Chen, ``{Large Language Models as Optimizers},'' \emph{arXiv preprint arXiv:2309.03409}, 2023.

\bibitem{Go2009-Twitter}
A.~Go, R.~Bhayani, and L.~Huang, ``{Twitter Sentiment Classification using Distant Supervision},'' \emph{CS224N project report, Stanford}, vol.~1, no.~12, p. 2009, 2009.

\bibitem{misra2023Sarcasm}
R.~Misra and P.~Arora, ``{Sarcasm Detection using News Headlines Dataset},'' \emph{AI Open}, vol.~4, pp. 13--18, 2023.

\bibitem{Sutskever2011generating}
I.~Sutskever, J.~Martens, and G.~E. Hinton, ``{Generating Text with Recurrent Neural Networks},'' in \emph{Proceedings of the 28th International Conference on Machine Learning}.\hskip 1em plus 0.5em minus 0.4em\relax Madison, WI, USA: Omnipress, 2011, pp. 1017--1024.

\bibitem{Ackley1985learning}
D.~H. Ackley, G.~E. Hinton, and T.~J. Sejnowski, ``{A learning algorithm for Boltzmann machines},'' \emph{Cognitive Science}, vol.~9, no.~1, pp. 147--169, 1985.

\bibitem{Holtzman2020TopP}
\BIBentryALTinterwordspacing
A.~Holtzman, J.~Buys, L.~Du, M.~Forbes, and Y.~Choi, ``The curious case of neural text degeneration,'' in \emph{International Conference on Learning Representations}, 2020. [Online]. Available: \url{https://openreview.net/forum?id=rygGQyrFvH}
\BIBentrySTDinterwordspacing

\bibitem{Schuller13-TI2}
B.~Schuller, S.~Steidl, A.~Batliner, A.~Vinciarelli, K.~Scherer, F.~Ringeval, M.~Chetouani, F.~Weninger, F.~Eyben, E.~Marchi, M.~Mortillaro, H.~Salamin, A.~Polychroniou, F.~Valente, and S.~Kim, ``{The INTERSPEECH 2013 Computational Paralinguistics Challenge: Social Signals, Conflict, Emotion, Autism},'' in \emph{{Proceedings INTERSPEECH}}, 2013, pp. 148--152.

\bibitem{Good94-PT}
P.~Good, \emph{{Permutation Tests: A Practical Guide to Resampling Methods for Testing Hypotheses}}.\hskip 1em plus 0.5em minus 0.4em\relax New York City, NY, USA: Springer, 1994.

\end{thebibliography}
